\pdfoutput=1
\documentclass[
]{ceurart}

\usepackage{listings}
\begin{document}

\copyrightyear{2022}
\copyrightclause{Copyright for this paper by its authors.
  Use permitted under Creative Commons License Attribution 4.0
  International (CC BY 4.0).}

\conference{PMAI@IJCAI22: International IJCAI Workshop on Process Management in the AI era, July 23, 2022, Vienna, Austria}

\title{Towards Automated Process Planning and Mining}
\title[mode=sub]{Mining process data for process improvements}
\title[mode=sub]{(Discussion/Short Paper)}

\author[1,2]{Peter Fettke}[%
orcid=0000-0002-0624-4431,
email=peter.fettke@dfki.de,
]
\cormark[1]

\author[1,2]{Alexander Rombach}[%
email={alexander_michael.rombach@dfki.de},
]

\address[1]{German Research Center for Artificial Intelligence (DFKI),
Saarbruecken, Germany}
\address[2]{Saarland University, Saarbruecken, Germany}
\cormark[1]

\cortext[1]{Corresponding author.}

\begin{abstract}
    AI Planning, Machine Learning and Process Mining have so far developed into separate research fields. At the same time, many interesting concepts and insights have been gained at the intersection of these areas in recent years. For example, the behavior of future processes is now comprehensively predicted with the aid of Machine Learning. For the practical application of these findings, however, it is also necessary not only to know the expected course, but also to give recommendations and hints for the achievement of goals, i.e. to carry out comprehensive process planning. At the same time, an adequate integration of the aforementioned research fields is still lacking. 
    In this article, we present a research project in which researchers from the AI and BPM field work jointly together. Therefore, we discuss the overall research problem, the relevant fields of research and our overall research framework to automatically derive process models from executional process data, derive subsequent planning problems and conduct automated planning in order to adaptively plan and execute business processes using real-time forecasts.
\end{abstract}


\maketitle

\section{Introduction}

Business Process Management (BPM) plays a vital role for organizations. The consistent design, implementation, execution and monitoring of business processes promises efficiency increases and cost reduction. In today’s organizations, BPM follows the paradigm of continuous process improvement. After a process is designed, implemented and executed, the process execution data is retrospectively analyzed with methods such as Process Mining in order to optimize process design in a cyclic manner. In this framework however, process agility (e.g. process variants) has to be defined at design time in order to achieve full process automation, which is typically not feasible in real-world scenarios.
Approaches using Process Mining and Machine Learning (ML) to overcome this problem exist, however stakeholders are left to their own on how to use these insights \cite{Neu}. At the same time, existing research regarding applications of AI planning in BPM focuses on the design phase and lacks planning of dynamic process executions. Such methods also require an extensive manual definition of the planning problem beforehand. To add on that, conventional Process Mining techniques are not necessarily suitable as a foundation for subsequent automated planning, since discovered process models purposely filter out information and/or simplify process details for the sake of readability. 

\begin{figure}
  \centering
  \includegraphics[width=\linewidth]{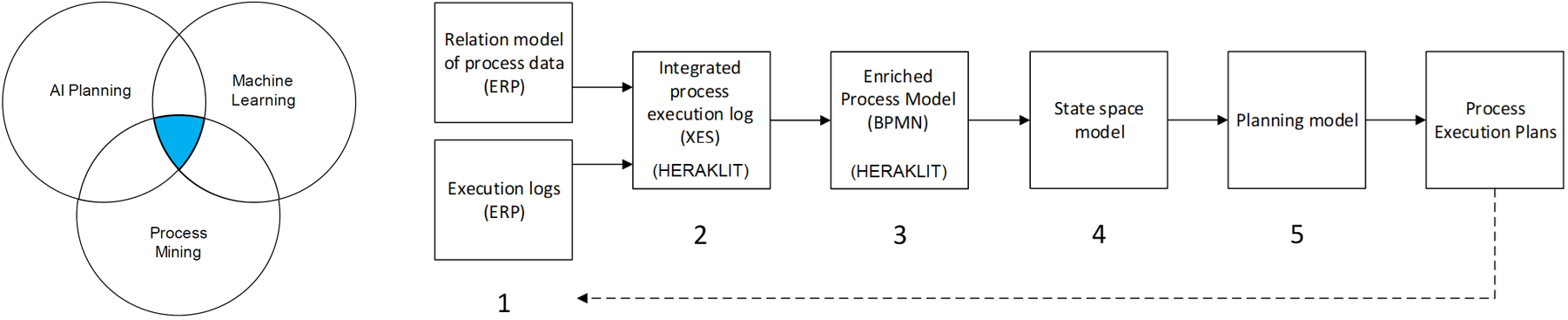}
  \caption{Research gap and overall research framework}
  \label{fig:combined}
\end{figure}

Our work is positioned within this research gap at the intersection of AI Planning, ML and Process Mining, as depicted in Figure \ref{fig:combined}. We propose developing new methods to achieve online automated planning leading to an agile execution of business processes. In that regard, we aim at answering the following research question: 
\textit{How is an automated planning of processes achieved using automatically derived process models and automatically constructed planning models?}

\section{Overall research framework}
\label{framework}

We propose a procedure model divided into five phases leading from executional process data to executable planning models, as depicted in the right part of Figure \ref{fig:combined}.
The general starting point is an ERP system providing process execution data in shape of relational data and execution logs. 
The second stage deals with extracting a process execution log covering the occurring cases and events. 
Based on that, the next important step is to automatically derive a process model. The key aspect hereby is that the resulting model entails enough process details to allow for AI planning, e.g. not only the general process flow, but also implicit domain knowledge, involved resources or potential resource allocation constraints. Such elements can then be used to automatically construct a sophisticated planning domain in subsequent phases. 
In this regard we also propose taking a broader view and starting to think about mining entire system models which describe the architecture of (hierarchical) systems including their static as well as dynamic elements based on distributed runs observable from event logs \cite{heraklit}.
Stage four covers the automated extraction of a state space model from the previously discovered model, which will then finally be (ideally automatically) transformed into a process planning model to conduct AI planning in stage five.
A plethora of different techniques can be integrated during this phase. One possibility is using ML-based methods like DeepQ learning, as it will be portrayed in section \ref{results}.
These automated conversions once again illustrate the before mentioned necessity of comprehensive process models, conventional Process Mining techniques usually cannot provide. 
The planning model is used to obtain an optimal execution of the conceptual process model. Resulting process plans are executed and thus generate new executional process data, leading to an iterative pipeline that ultimately allows to conduct automated process planning.

\section{Application cases}
\label{cases}

\begin{figure}
  \centering
  \includegraphics[width=0.8\linewidth]{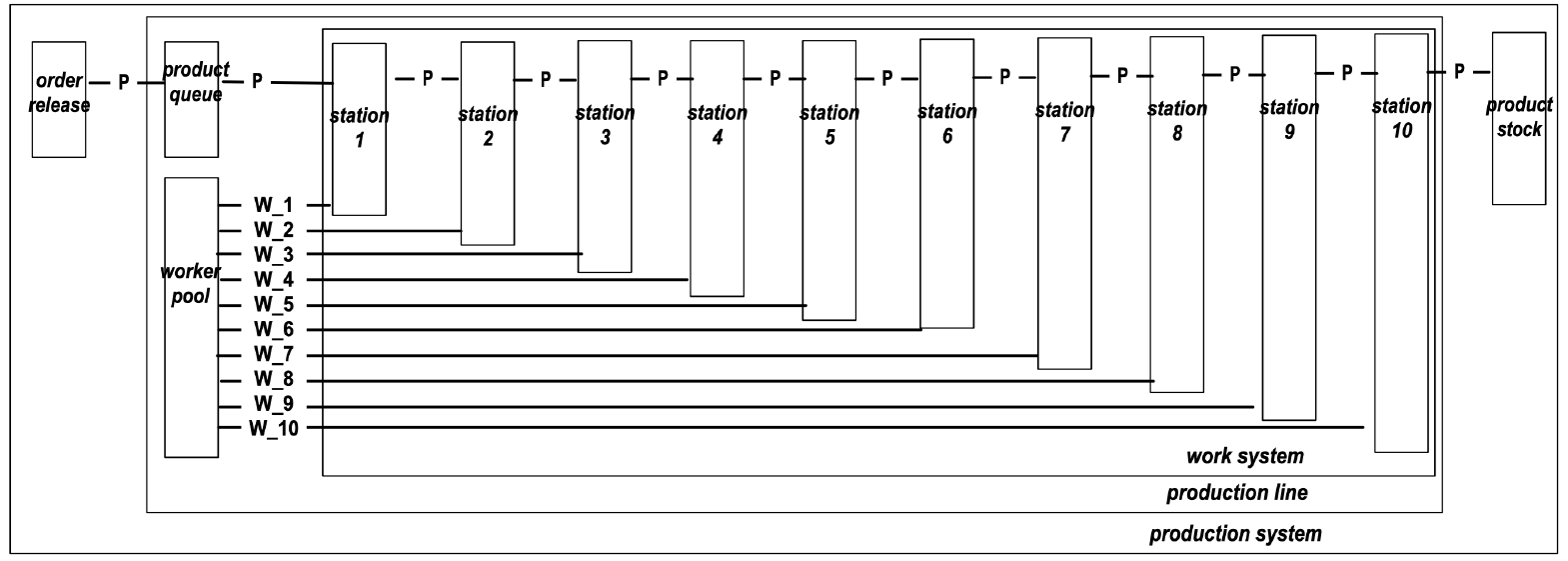}
  \caption{Abstract view of an exemplary application case}
  \label{fig:bosch}
\end{figure}

We designed different application cases to develop and evaluate our approach, most prominently a classic flow-shop scheduling problem shown in Figure \ref{fig:bosch} modeled using \textsc{Heraklit} \cite{heraklit}. 

In this scenario, a set of different products have to be produced on a production line consisting of different machine stations and workers to operate them, where each product has to cross all stations. Figure \ref{fig:bosch} shows an exemplary setup consisting of 10 stations and 10 workers. The scenario includes several variable factors, e.g. the process execution time of one production step depends on the underlying product, station as well as the operating worker with a distinct experience level. The general steps to perform at each station can differ, adding another dimension to the planning problem. On top of that, the process includes walking times from one station to another whenever necessary, which in turn is dependent on the layout of the production line, i.e. how the individual stations are distributed.

\section{Preliminary results}
\label{results}

Regarding phases two and three, multiple modeling frameworks can be considered. While XES is a well-known standard for event logs, we came to the conclusion that such representation might not suffice due to not providing enough information about involved resources necessary for planning domains. \textsc{Heraklit} on the other hand is a modeling language based on symbolic Petri Nets which we regard as a good foundation for subsequent planning procedures. It also allows us to model entire system models as described in section \ref{framework} including dependencies between individual subsystems and agents. For more details about \textsc{Heraklit}, we refer to \cite{heraklit}.

Several advances have been made regarding the latter parts of the framework, e.g. “domain-specific Constraint Patterns” in the context of Constraint Programming \cite{patterns}. These patterns describe reoccurring problems and their solutions and can be combined in order to simplify the construction of planning problems. An example of such pattern is the flow-shop. It is also possible to automatically identify occurrences of these patterns in event logs and automatically assemble parameterized constraint models. This assembly is ultimately based on a pattern repository, which therefore requires manual yet reusable one-time definitions of these patterns. 

We also propose a novel GPU-assisted approach for on-the-fly generation of training samples for Deep-Q learning \cite{macsq}, a powerful solution for the main challenge of providing training data for such techniques, which outperforms conventional approaches both in terms of run-time and memory/storage consumption. It uses the domain description extracted from a process model in order to iteratively instantiate randomized states and compute corresponding Q-matrices. As a Proof of Concept, the input models are preliminarily provided in PNML file formats. In the future we plan on integrating richer representations like \textsc{heraklit} in order to increase the amount of usable domain knowledge, which is necessary in more complex use cases. 

Based on this approach, we developed Deep-Q-Agents methods for optimization. To achieve convincing results using such techniques, it is necessary to set up domain specific configurations, which in turn requires domain knowledge. This however also requires a human assessment of these configurations. To overcome this problem, we suggest an automated exploration of the underlying search space as a heuristic to find potentially optimal configurations. 

\section{Conclusion and outlook}

Several research initiatives in the fields of AI Planning and Process Mining in the context of BPM have been made so far. However, there are no satisfying results when it comes to planning and agile executions of business processes in an automated fashion. 
To this end, we propose an iterative procedure model build upon executional process data to automatically derive process models and automatically infer solvable planning problems to plan and optimize business processes. Up until now, several concepts have been implemented and advancements addressing individual parts of the procedure model have been made. The next main step is to adequately combine these individual concepts and put the proposed "round-trip" into practice. 

\begin{acknowledgments}
    This work is part of the research project APPaM (Grant 01IW20006),
    which is partly funded by the Federal Ministry of Education and Research (BMBF).
\end{acknowledgments}


\bibliography{bib}

\end{document}